\documentclass[journal]{IEEEtran}
\usepackage[caption=false]{subfig}
\usepackage{multirow}
\usepackage{amsmath}
\usepackage{amssymb}
\usepackage{lipsum}
\usepackage{cuted}
\usepackage{picinpar, float, graphicx}
\usepackage{booktabs}
\usepackage{array}
\usepackage{longtable} 
\usepackage{epstopdf}
\usepackage{setspace} 
\usepackage{xcolor}
\usepackage{subfig}
\usepackage{gensymb}
\usepackage{soul} 
\usepackage[normalem]{ulem}
\usepackage{bm}
\usepackage{graphicx}
\usepackage{mathrsfs} 
\usepackage{algorithm}
\usepackage{algorithmic}
\usepackage{threeparttable} %
\usepackage{relsize}
\usepackage{textcomp,gensymb}
\usepackage{tikz,xcolor,hyperref}

\definecolor{green}{RGB}{0,100,0}

\definecolor{lime}{HTML}{A6CE39}
\DeclareRobustCommand{\orcidicon}{
\begin{tikzpicture}
\draw[lime, fill=lime] (0,0)
circle[radius=0.16]
node[white]{{\fontfamily{qag}\selectfont \tiny \.{I}D}}; 
\end{tikzpicture}
\hspace{-2mm}
}
\foreach \x in {A, ..., Z}{%
\expandafter\xdef\csname orcid\x\endcsname{\noexpand\href{https://orcid.org/\csname orcidauthor\x\endcsname}{\noexpand\orcidicon}}
}

\begin{document}

\title{Cable-driven Continuum Robotics: Proprioception via Proximal-integrated Force Sensing}

\author{ Gang Zhang\orcidB{}, Junyan Yan\orcidA{}, Jibiao Chen\orcidC{}, and Shing Shin Cheng\orcidD{} 
\thanks{This work was supported in part by National Key Research and Development Program of China (2023YFE0203200), in part by Innovation and Technology Commission of Hong Kong (ITS/235/22, ITS/225/23, ITS/224/23, MHP/096/22 and Multi-scale Medical Robotics Center (InnoHK Initiative)), and in part by Research Grants Council of Hong Kong (CUHK 14217822, CUHK 14207823, CUHK 14211425, and AoE/E-407/24-N), and in part by Curve Robotics Limited, Hong Kong. (Corresponding authors: Junyan Yan and Shing Shin Cheng)

\  Gang Zhang, Junyan Yan, and Jibiao Chen are with the Department of Mechanical and Automation Engineering and T Stone Robotics Institute, The Chinese University of Hong Kong, Hong Kong, and also with Curve Robotics Limited, Hong Kong (Email: gangzhang@link.cuhk.edu.hk, junyanyan@mae.cuhk.edu.hk, jibiaochen@link.cuhk.edu.hk).

\ Shing Shin Cheng is with the Department of Mechanical and Automation Engineering and T Stone Robotics Institute, The Chinese University of Hong Kong, Hong Kong, also with the Shun Hing Institute of Advanced Engineering and Multi-Scale Medical Robotics Center, Hong Kong, and also with Curve Robotics Limited, Hong Kong (Email: sscheng@cuhk.edu.hk).}

\thanks{This work has been submitted to the IEEE for possible publication. Copyright may be transferred without notice, after which this version may no longer be accessible.}
}
\markboth{IEEE Transactions on ,~Vol.~X, No.~X, August~2024}%
{Shell \MakeLowercase{\textit{et al.}}: A Sample Article Using IEEEtran.cls for IEEE Journals}


\maketitle

\begin{abstract}
Micro-scale continuum robots face significant limitations in achieving three-dimensional contact force perception, primarily due to structural miniaturization, nonlinear mechanical, and sensor integration. To overcome these limitations, this paper introduces a novel proprioception method for cable-driven continuum robots based on proximal-integrated force sensing (i.e., cable tension and six-axis force/torque (F/T) sensor), inspired by the tendon-joint collaborative sensing mechanism of the finger. By integrating biomechanically inspired design principles with nonlinear modeling, the proposed method addresses the challenge of force perception (including the three-dimensional contact force and the location of the contact point)  and shape estimation in micro-scale continuum robots. First, a quasi-bionic mapping between human tissues/organs and robot components is established, enabling the transfer of the integrated sensing strategy of tendons, joints, and neural feedback to the robotic system. Second, a multimodal perception strategy is developed based on the structural constraints inherent to continuum robots. The complex relationships among mechanical and material nonlinearities, robot motion states, and contact forces are formulated as an optimization problem to reduce the perception complexity. Finally, experimental validation demonstrates the effectiveness of the proposed method.
This work lays the foundation for developing safer and smarter continuum robots, enabling broader clinical adoption in complex environments.

\end{abstract}


\section{Introduction}
\subsection{State-of-the-Art Review}

Cable-driven mechanisms are the most classic actuation method for continuum robots~\cite{zhou2024submillimeter}. Compared to other types of continuum robots, cable-driven designs exhibit clear bio-inspired principles~\cite{li2024design,russo2023continuum}. Analogous to the human body, the cables act as muscles, the backbone as the skeleton, and the computer as the brain. Through coordinated interactions between the cables and backbone under computer control, continuum robots can execute complex motions in confined spaces. These unique capabilities make them increasingly valuable in medical fields, including neurosurgery~\cite{kim2017toward,qi2025development}, otolaryngology~\cite{xu2024novel,hong2018development}, and laparoscopic surgery~\cite{wu2019design,wu2022robotic}.

Although the structural similarities between continuum robots and human bodies are striking, achieving human-like proprioception in continuum robots remains a critical and challenging research problem. Proprioception refers to a system’s ability to perceive its own state, encompassing position, posture, motion, and contact forces~\cite{tuthill2018proprioception}. Analogous to the human body, proprioception in continuum robots can be broadly categorized into two aspects: shape estimation~\cite{liu2022morphology} and contact force perception~\cite{yao2023rnn}. While contact perception is often considered a type of proprioception~\cite{dupont2022continuum}, in this context, it is treated as a subset of contact force perception. 

\begin{table*}\footnotesize
\begin{threeparttable} 
    \centering
    \caption{Current status of research on continuum robot sensing}
   \setlength{\tabcolsep}{.0038\linewidth}
    \begin{tabular}{ccccccccccll}
    \hline
References &  Categories &Input
& Output& Sensor &location& Model& Contact&  Friction&Dimension& Accuracy&Speed\\
 \hline
 Li et al~\cite{li2024wavelength}&Sensor-based& -& Force& FBG& Body& -& Tip& -& 3& 6\%&-\\
 Yang~\cite{zhao2024step}& Sensor-based& -& Force& TS& Body& -& Local& -& 3& 4.4g&2.5 Hz\\
 Zhao et al.~\cite{zhao2024step}& Sensor-based& -& Force& FBG& Tip& -& Local& -& 3& 3.46\%&5000 Hz\\
 Luo et al.~\cite{luo2025embedded}& Sensor-based&-& Shape& OS& Body& -& Local& -& 1& -&100 Hz\\
Gao et al.~\cite{gao2024body}&  Sensor-based&-&Shape+Force& FBG &Body& -& Any&  -&3& 0.31g&-\\
Yao et al.~\cite{yao2023rnn}&  Sensor-based&-&Force& OS &Tip& -& Tip&  -&3& 15.5g&-\\
 Xu et al.~\cite{xu2008investigation}&  Model-based&Internal & Shape+Force& - &-& CC & Tip&  -&1& 0.34±0.83g&-\\
 Leung et al.~\cite{leung2024hybrid}&  Model-based&Internal& Shape+Force& - &-& Observer& Tip&  -&1& 1.46g&-\\
 Xiang et al.~\cite{xiang2023learning}&  Model-based&Internal& Shape+Force& - &-& LSTM& Tip&  -&1& 0.24g&1.88 Hz\\
Hao et al.~\cite{hao2023two} & Model-based& Internal&Shape+Force& -& -& LSTM+MLP& Tip& -& 1& 26g&-\\
Du et al.~\cite{du2024sensor}&  Model-based&Internal&Shape+Force& - &-& BCM & Tip&  Y&1& 0.74g&-\\
Zhang et al.~\cite{zhang2024design}&  Model-based&Internal&Shape+Force& - &-& BCM & Tip&  Y&1& 1.95g&-\\
Feliu-Talegonet al.~\cite{feliu2025actuation}& Model-based& Internal& Shape+Force& -& -& Cosserat& Tip& -& 1& 0.46g&6 Hz\\
Wang et al.~\cite{wang2024sensing}&  Hybrid&Internal+Strain & Shape+Force& PSS &Body& PCC & Tip&  -&3& 60g&2.5 Hz\\

Aloi et al.~\cite{aloi2022estimating}&  Hybrid&Internal+Positon& Shape+Force& Vision &External& Cosserat & Any&  -&1& 58g&0.47 Hz\\
Thuruthel et al.~\cite{thuruthel2019soft}&  Hybrid&Internal+Strain& Shape+Force& cPDMS &Body& LSTM& Tip&  -&1& 5±6g &10 Hz\\
Wang et al.~\cite{wang2024estimating}&  Hybrid&Internal+Strain& Shape+Force& FBG &Body& Cosserat & Any&  Y&1& 2.4g&-\\
Xu et al.~\cite{xu2016curvature}&  Hybrid&Internal+Strain& Shape+Force& FBG &Any& Cosserat& Tip&  -&1& 5\%&100 Hz\\
Khan et al.~\cite{khan2017force}&  Hybrid&Internal+Strain& Shape+Force& FBG &Body& Cosserat& Tip&  -&3& 1.59g&-
\\
Li et al.~\cite{li2025collaborative}&  Hybrid&Internal+position & Shape& US-EM &Tip& Bezier curve & -&  -&-&-&7 Hz\\
Lilge et al.~\cite{lilge2022continuum}& Hybrid& Internal+Pose/Strain&Shape & EM& Body& Cosserat& -& -& -& -&100 Hz\\
Lilge et al.~\cite{lilge2024state}& Hybrid& Internal+Strain& Shape& FBG& Body& Cosserat& -& -& -& -&100 Hz\\
Ha et al.~\cite{ha2022shape}& Hybrid& Internal+Strain&Shape  & FBG& Body& ANN& -& -& -& -&40 Hz\\
Zheng et al.~\cite{zheng2025estimatingdynamicsoftcontinuum}& Hybrid& Internal+F/T&Shape  & F/T& Base& Cosserat& -& -& -& -&40 Hz\\
This paper&  Hybrid& Internal+F/T&Shape+Force& F/T &Base& BCM & Any&  Y&3& 0.93g&100 Hz\\ \hline
    \end{tabular}
 \label{Current_status}
{ \begin{tablenotes}[flushleft]\footnotesize
 \item{"Internal" refers to actuation information such as cable displacement, cable tension, rotation angle, etc. "Strain" refers to the deformation of the sensor. "Position"  refers to the position of points on the robot. "Pose" refers to the posture of the robot tip.
“PSS”, "TS", “OS”, “cPDMS”, "US", "EM", “F/T”, and “FBG” refer to piecewise soft sensor, temperature sensor, optical switches, polydimethylsiloxane impregnated with conductive carbon nanotubes, ultrasound, electromagnetic, six-axis force/torque sensor, and fiber bragg grating, respectively.
“PCC”, “EBB”, and “BCM” refer to piecewise constant curvature, Euler–Bernoulli beam, and beam constraint model, respectively.
“SVR”, "ANN", and “LSTM” refer to support vector regression, artificial neural networks, and long short-term memory, respectively.}

\end{tablenotes}}
\end{threeparttable}
\vspace{-2 em}
\end{table*}

Various strategies have been explored to achieve proprioception in continuum robots, which can be broadly categorized into sensor-based methods (using sensor measurements), model-based methods (using internal-driven inputs), and hybrid methods (combining both), as summarized in Table~\ref{Current_status}. Sensor-based methods typically integrate strain gauges, fiber optic sensors, or piecewise soft sensors into the robot to directly capture deformation or force information~\cite{gao2023progress}. These approaches offer accurate and real-time measurement of both shape and contact forces. However, they require additional space for sensor integration, which poses a significant challenge for small-scale continuum robots. In particular, realizing simultaneous force and shape sensing in such compact designs remains extremely difficult~\cite{li2024wavelength,zhao2024step}.

Model-based approaches treat the entire continuum robot as a self-sensing system, enabling force and shape estimation based on internal actuation information (cable displacement and internal cable tensions). 
Early work by Xu et al.~\cite{xu2008investigation} attempted to predict the tip force of a continuum robot using cable tension data within the constant curvature (CC) model and contact direction. More recently, Du et al.~\cite{du2024sensor} employed the beam constraint model (BCM) to estimate contact forces, achieving more accurate measurement of unidirectional tip forces with known contact location and direction. 
Learning-based approaches have recently become a major research focus, offering a data-driven alternative to explicit modeling. Instead of relying on analytical formulations, these methods leverage neural networks to capture the nonlinear relationship between actuation signals and contact forces. For instance, Xiang et al.~\cite{xiang2023learning} employed a long short-term memory (LSTM) neural network to achieve contact force perception when the contact force orientation and contact location are known. Hao et al.~\cite{hao2023two} further advanced this by using a multilayer perceptron (MLP) and LSTM to predict unidirectional tip forces at the distal end of the robot. 
The latest model-based approach is to build observers to achieve proprioception. 
Leung et al.~\cite{leung2024hybrid} developed an observer based on past experience to estimate the tip contact force. The observer estimates the contact force based on the configuration and driving information of the soft segment.

As a result, their applicability is often restricted to simplified scenarios with limited generalizability, such as quasi-static conditions with known contact location and orientation. In more complex situations involving interactions with unknown contact location and contact force orientation, or highly uncertain anatomical environments, these methods are no longer applicable, which limits their practical use in real surgical or unstructured environments.

Hybrid approaches combine sensor feedback with physical modeling to reduce system identification complexity and overcome dimensional constraints. A representative strategy within hybrid approaches is to employ shape sensing as an intermediary for force estimation. For example, Aloi et al.~\cite{aloi2022estimating} reconstructed the continuum robot’s shape through visual tracking and applied an extended Kalman filter to analyze discrepancies between modeled and measured geometries under distributed loading, thereby enabling multidimensional contact force perception. Wang et al.~\cite{wang2024sensing} followed a similar idea by introducing a piecewise soft sensor, where contact forces were inferred from discrepancies between the sensed shape and the predictions of a piecewise constant curvature (PCC) model. In recent years, hybrid approaches have been developed as state estimation frameworks that combine continuum mechanics and sensor information for stable perception. Lilge et al.~\cite{lilge2022continuum,lilge2024state} demonstrated this by reconstructing the complete robot state from limited discrete sensing from the electromagnetic sensors using a simplified Cosserat rod model.
{Recent advances in boundary observers for continuum robots, notably the tip-velocity-based estimator proposed by Zheng et al. ~\cite{zheng2025estimatingdynamicsoftcontinuum}, have demonstrated the feasibility of recovering full robot states using minimal sensing. While inspiring, their work focuses on dynamic state estimation under known inputs without external disturbances. In contrast, this paper addresses the more challenging problem of simultaneous shape, 3D contact force, and contact location estimation under unknown external interactions—a scenario inherent to physical interaction tasks. }

\subsection{Challenges and Contributions}

Proprioception in continuum robots has primarily focused on contact force perception, contact location perception, and shape estimation. Despite notable progress, a major challenge persists: acquiring richer perceptual information with minimal sensor usage, particularly in compact medical environments where space is highly constrained. In medical robotics, the ability to sense three-dimensional forces at uncertain contact locations on instruments under 4 mm in diameter remains a longstanding and unresolved bottleneck.

Most prior work has concentrated on low-dimensional force sensing (e.g.,  one-dimensional gripping forces, or transverse contact force on the robot backbone) within miniaturization constraints, or has simplified the problem by assuming fixed contact location. 
Sensor-based approaches enable direct and fast measurements for real-time response but are limited by sensor bulk, integration complexity, and trade-offs among cost, accuracy, and stability. Model-based approaches instead infer forces and shape from internal (actuation) inputs such as cable tension or displacement, often through mechanical or data-driven models. However, since they typically rely on assumptions about unknowns such as contact location or contact force orientation, the perceptual information they deliver is limited. In essence, these approaches attempt to resolve inherently underdetermined systems, which can be alleviated by introducing additional assumptions or sensor feedback. To overcome this limitation, hybrid approaches have been proposed, aiming to combine the complementary strengths of sensors and models to enhance the perception capabilities of continuum robots. Nevertheless, significant challenges persist, including inaccuracies in physical models, the reliability of hybrid strategies, and the practical constraints associated with integrating additional sensors in space-limited clinical environments.

Motivated by these challenges and inspired by biological proprioception, this paper introduces a framework that integrates force information at the robot's proximal end with a high-fidelity mechanical model. By combining the strengths of sensor-based and model-based methods, the framework achieves simultaneous contact force, contact location, and shape estimation while avoiding the spatial limitations that constrain sensor placement in compact continuum robot structures. This strategy provides an efficient and practical solution for proprioception in complex, size-constrained environments, especially in medical robotics. The contributions of this paper are as follows:
{
\begin{enumerate} 
{
\item {Bio-Inspired Sensing Framework with Proximal-Only Sensing: Implements a bio-inspired sensing paradigm using a proximal six-axis force/torque sensor and cable tension feedback to mimic human joint-tendon mechanisms, enabling submillimeter shape perception, three-dimensional force estimation with sub-gram accuracy, and millimeter-level contact location estimation.
}
\begin{itemize} 
\item Inspired by the proprioceptive mechanisms of tendons and joints in the human physiological structure, our method integrates a proximal six-axis force/torque sensor to emulate the role of joint proprioceptors and a cable tension feedback sensor to mimic tendon organ feedback. By integrating these signals through nonlinear modeling, the robot achieves accurate 3D contact force perception and shape estimation at the microscale.
\item {Experimental validation shows that this method can achieve an $\leq$1 mm average tip position error, a 0.93 g average force estimation error, and a contact location error of $\leq$3 mm.}
\end{itemize} 

\item {Duality-Based Optimized Perception for Force-Location Co-Estimation: Develops an optimization method that leverages the duality between cable tension and base force, reformulating the coupling of mechanical nonlinearity, material properties, motion states, contact location, and contact forces into a unified optimization problem to directly compute contact force magnitude and orientation.}
\begin{itemize} 
\item The internal cable displacement serves as the boundary condition, while cable tension is incorporated to resolve the under-determination of the perception model. The magnitude of the contact force is directly calculated from the duality between cable tension and base force information, and subsequently used in the optimization framework to determine the contact location.

\item Experiments show that the strategy can stably perceive the contact force under different states (dynamic friction, acceleration, and deceleration). 
\end{itemize} 
\item {Friction-Uncertainty Mitigation for Stable Force Perception: Introduces active/passive contact classification to normalize uncertain internal friction, enabling stable and accurate force and position sensing during physical interaction.}

\begin{itemize} 
\item {Two contact modes, active and passive, are proposed to accurately classify contact states, as shown in Fig.~\ref{Main}(b), thereby achieving more stable and accurate multimodal contact force sensing.}

\item Inspired by the human strategy of repeatedly grasping an object to estimate its weight, the robot performs a reciprocating motion under load to compensate for internal friction uncertainties arising from unknown contact interactions.

\item Experiments have shown that this method can achieve accurate perception of multi-dimensional force size and position, with an average contact location error of $\leq0.63$ mm and an average multi-dimensional force error of $\leq2.20$ g. The proposed strategy can reduce the average contact force position error from 1.67 mm to 0.21 mm.
\end{itemize} 
}
\end{enumerate}}

The rest of this paper is organized as follows. Section II introduces the principles of perception and the modeling approach. Section III designs experiments to verify the proposed model. Section IV discusses the methodology and experimental results. Section V concludes the paper.

\section{Methods }

\subsection{Basic Principles} 
\subsubsection{Bionic principles}

 \begin{figure*}[htpb]
\centering  
\includegraphics[width=180mm]{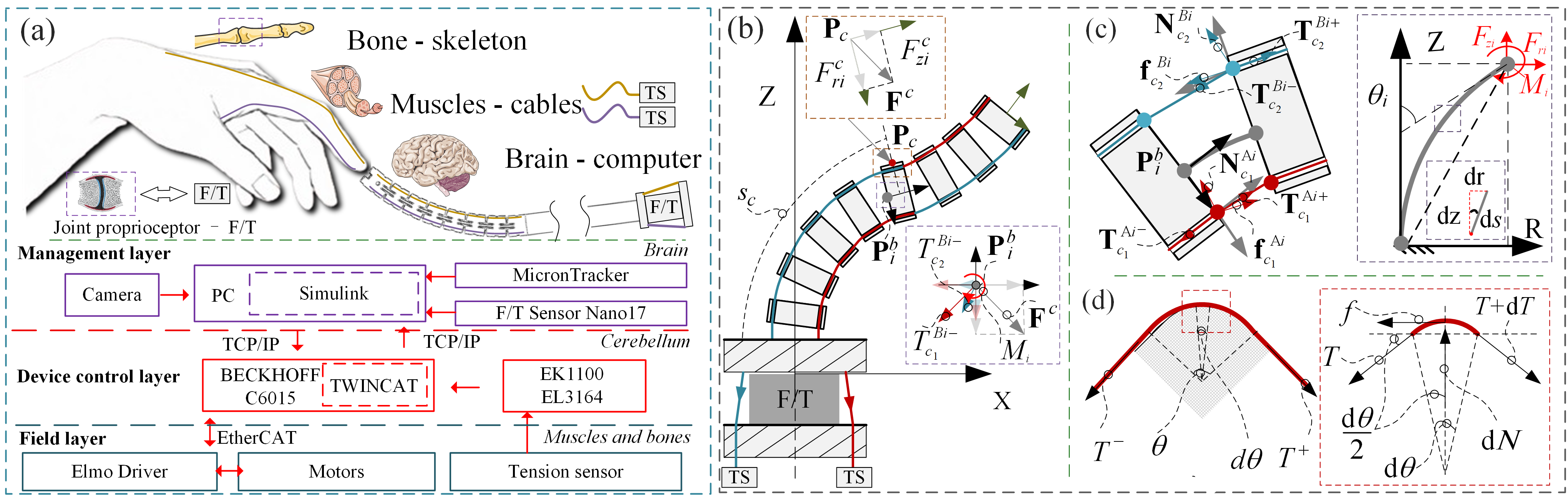}
\caption{Basic principles of bionics and system framework. (a) Biological mapping of robotic sensing components.
(b) Principle of implementing proprioception. (c) Mechanical balance between cable and beam. (d) Capstan friction model to describe cable tension transmission.
}
\label{Main}   
\vspace{-15 pt}
\end{figure*}

{Joint proprioceptors in organisms, including muscle spindles that detect changes in muscle length and tendon organs that sense muscle tension, as well as joint capsule receptors such as Ruffini endings, Pacinian corpuscles, Golgi tendon organ-like endings, and free nerve endings that collectively respond to forces, torques, joint angles and velocities, enable integrated perception and regulation of contact forces, joint positions, and self-motion by transducing mechanical stimuli through sensory nerve endings to the central nervous system~\cite{riemann2002sensorimotor}, as illustrated in Fig.~\ref{Main}.} Inspired by this, this paper installs an F/T sensor at the proximal end of the robot to simulate the function of joint proprioceptors, enabling real-time monitoring of the robot's overall force state. Simultaneously, tension sensors are placed on the drive cables to simulate the stretch-sensing capabilities of muscles and measure the local force state of individual cables. Through this collaborative design of joint-level (global) and muscle-level (local) sensing, the robot is able to comprehensively perceive its own form and contact forces. 
Inspired by the distinction in biological proprioception between externally induced passive contact and self-generated active contact, the robot’s sensing is similarly categorized into active and passive contact modes, allowing coordination of multiple sensors and the establishment of a comprehensive perception framework. Based on this framework, combined with quasi-static mechanical modeling of cable-driven systems, the robot can accurately infer contact force and interaction states from proximal sensing information.
\subsubsection{Basic Mathematical Principles} 
The contact point is denoted as ${\mathbf{P}}_c (x_c, y_c, z_c)$. By combining the governing equations, an implicit equations system $\mathbf{F}(\mathbf{x})$ with six unknowns is obtained: 
\begin{small}
\begin{equation}
\left\{ \begin{array}{l}
\mathbf{x}(t)
=(x_c(t), y_c(t), z_c(t), \textbf{F}^C(t))^T\in\mathbb{R}^6, \\
\mathbf{F}(\mathbf{x})={\bf 0},\\
\end{array} \right.
\label{F0}
\end{equation}
\end{small}
where ${\bf{F}}^{C}$ is the contact force, $t$ represents the current time.
Solving these nonlinear equations forms the basis of force sensing in robots. However, the intrinsic coupling between contact force and shape estimation complicates the process. Shape estimation involves variables—the beam deformations—which are directly coupled with the contact forces. This coupled relationship can be formulated within the framework of elasticity mechanical theory, which consists of the fundamental components: proximal forces decoupling, mechanical equilibrium, beam deformation, and geometric constraints. 
{
\subsection{Proximal Forces Decoupling}
The principle of using the robot's proximal end sensors' information to sense the contact force and the position of the contact point is shown in Fig.~\ref{Main}(b). Separates the cable tension from the contact forces to determine the magnitude of the contact forces using the following equation, the number of unknowns in Eq.~\eqref{F0}. can be reduced to 3. 
\begin{equation}
{\textbf{F}}^C(t)={\textbf{F}}^M(t)-{\bf{F}}_{C_a}^{in}(t),
\label{F1}
\end{equation}
where ${\textbf{F}}^M$, ${\bf{F}}_{C_a}^{in}$ are forces measured by the F/T sensor at the proximal end of the robot, and the input cable forces. It should be noted that the contact force obtained is expressed in the global coordinate system. For systems comprising multiple deformable elements, such as continuum robots, determining the contact position through shape calculation requires transforming these forces into the local coordinate system of each individual deformable element (e.g., beam).}
\subsection{Mechanical Balance} 
{The purpose of the mechanical balance is to determine the forces acting on the tip of each beam in local coordinate systems, as illustrated in Fig.~\ref{Main}(c).} Assume that the direction of the support force is on the angle bisector of the tension forces on both sides~\cite{du2024sensor}. The relationship between the cable tensions and the forces at the joint $i$ can be expressed as: 
\begin{equation}
\left\{ {\begin{array}{*{20}{c}}
{ - T_{{C_a}}^{{A_i} - }\cos \left( {\frac{{\theta_i }}{2}} \right) - u_{C_a}^i{N_i} + T_{{C_a}}^{{A_i} + }\cos \left( {\frac{{\theta_i }}{2}} \right) = 0},\\
{ - T_{{C_a}}^{{A_i} - }\sin \left( {\frac{{\theta_i }}{2}} \right) +{N_i} + T_{{C_a}}^{{A_i} + }\sin \left( {\frac{{\theta_i }}{2}} \right) = 0},
\end{array}} \right.
\label{M1}
\end{equation}
where $A_i$ represents the contact point near the beam base, $C_a$ represents the cable, $i$ is the number of the joint, and $a$ is the number of the cable. According to the capstan friction theory~\cite{yang2024novel}, as shown in Fig.~\ref{Main}(d), the cable tension relationship between the contact point can be expressed by:
\begin{equation}
\left\{ \begin{array}{l}
T_{{C_a}}^{{A_i} + } = {e^{u_{C_a}^i\varphi _{{C_a}}^{{A_i}}}}T_{{C_a}}^{{A_i} - },\\
T_{{C_a}}^{{B_i} - } = T_{{C_a}}^{{A_i} + },\\
T_{{C_a}}^{{B_i} + } = {e^{u_{C_a}^i\varphi _{{C_a}}^{{B_i}}}}T_{{C_a}}^{{B_i} - },
\end{array} \right.
\label{M2}
\end{equation}
where $B_i$ represents the contact point near the beam tip. $u_a^i$ is the friction coefficient.{Based on the mutual movement trend of the cables within the joint, the quasi-static friction coefficient $u_a^i$ can be expressed by:} 
\begin{equation}
\left\{ \begin{array}{l}
u_{C_a}^i (t)= u,l_{C_a}^i(t) < l_{C_a}^i(t - 1),\\
u_{C_a}^i(t) =  - u,l_{C_a}^i(t) > l_{C_a}^i(t - 1),\\
 - u \le u_a^i (t)\le u,l_{C_a}^i(t) = l_{C_a}^i(t - 1),\\
 u_{C_a}^i(t) =  0,\textit{slack}.
\end{array} \right.
\label{M3}
\end{equation}
where $l$ is the length of the cable between the contact points. {$\mu$ is the static friction coefficient, and its measurement procedure can be found in~\cite{du2024sensor}.} $l_{C_a}^i(t)$ can be calculated using:
\begin{equation}
l_{C_a}^i(t)=\sqrt{(r_i^{B_{C_a}}(t))^2+(z_i^{B_{C_a}}(t))^2},
\label{M3_1}
\end{equation}
where $r_i^{B_{C_a}}$ and $z_i^{B_{C_a}}$ are the position of point ${\textbf{B}}$, which can express by:
\begin{small} 
\begin{equation}
\left[ {\begin{array}{*{20}{c}}
{r_i^{{B_{C_a}}}(t)}\\
{z_i^{B_{C_a}}(t)}
\end{array}} \right] = \left[ {\begin{array}{*{20}{c}}
{{r_i}(t)}\\
{{z_i}(t)}
\end{array}} \right] + \left[ {\begin{array}{*{20}{c}}
{{\mathop{\rm s}\nolimits} {\theta _i(t)}}&{{\mathop{\pm \rm c}\nolimits} {\theta _i(t)}}\\
{{\mathop{\rm c}\nolimits} {\theta _i(t)}}&{{\mp \mathop{\rm s}\nolimits} {\theta _i(t)}}
\end{array}} \right]\left[ {\begin{array}{*{20}{c}}
0\\
{{d_c}/2}
\end{array}} \right],\\
\label{M5}
\end{equation}
\end{small}
where $r_i(t)$ and $z_i(t)$ are the tip displacements of the beam $i$ in the local coordinate system. ${{\mathop{\rm s}\nolimits} {\theta _i(t)}}$ and ${{\mathop{\rm c}\nolimits} {\theta _i(t)}}$ are the abbreviations of ${{\mathop{\rm sin}\nolimits} {\theta _i(t)}}$ and ${{\mathop{\rm cos}\nolimits} {\theta _i(t)}}$.
In Eq.~\eqref{M2}, $\varphi _{{C_a}}^{{A_i}}(t) $ and $\varphi _{{C_a}}^{{B_i}} (t)$ are the deflection angles of the cable at points \textbf{A} and \textbf{B}, which can be calculated using:
\begin{equation}
\left\{ \begin{array}{l}
\varphi _{{C_a}}^{{A_i}}(t) = {\rm{atan}}(\frac{{r_i^{B_{C_a}}(t) - \frac{{{d_c}}}{2}}}{{z_i^{B_{C_a}} }(t)}),\\
\varphi _{{C_a}}^{{{\rm{B}}_i}}(t) = {\theta _i(t)} - \varphi _{{C_a}}^{{A_i}}(t).
\end{array} \right.
\label{M4}
\end{equation}

The subsequent joints are considered as a whole, the force exerted on the beam $i$ by the cable can be calculated using:
\begin{equation}
T_{{C_a}}^{{B_i} - }(t) = F_{{C_a}}^{in}(t) {e^{u_{C_a}^i(t) \varphi _{{C_a}}^{{A_i}}(t) \sum\limits_{i = 1}^{i - 1} \varphi _{{C_a}}^{{{\rm{B}}_i}}(t)u_{C_a}^i(t) }},
\label{M6}
\end{equation}
where $F_{{C_a}}^{in}$ is the force input by the cable $a$ at the proximal end. Set $F_{zi}$, $F_{ri}$ are the forces acting on the beam $i$ in the local coordinate system. 
\begin{small} 
\begin{equation}
\left\{ \begin{array}{l}
{F_{ri}}(t) = \sum\limits_{a = 1}^{{2}} {\left( {T_{{C_a}}^{{B_i} - }(t)s \varphi _{{C_a}}^{{A_i}}}(t) \right) + F_{ri}^C}(t), \\
{F_{zi}} (t)= \sum\limits_{a = 1}^{{2}} {\left( {T_{{C_a}}^{{B_i} - }(t)c \varphi _{{C_a}}^{{A_i}}(t)} \right) + F_{zi}^C} (t),\\
{\textbf{M}_i}(t) = (\sum\limits_{a = 1}^{{2}} { \left( {\textbf{D}_{C_a}\times \textbf{T}_{{C_a}}^{{B_i} - }(t)} \right) }
+ \textbf{M}_i^C(t)),
\end{array} \right.
\label{M7}
\end{equation}
\end{small} 

where $F_{xi}^C(t)$ and $F_{zi}^C(t)$ are the components of the external contact force ${{\bf{F}}^c_i}(t)$ in each direction in the local coordinate system. $\textbf{M}_i=[{M}_{xi},{M}_{yi},{M}_{zi}]$.
$\textbf{D}_{C_a}$ is the vector from point B to the beam tip in the local coordinate system. $\textbf{T}_{{C_a}}^{{B_i} - }(t)=[ {T_{{C_a}}^{{B_i} - }(t)s \varphi _{{C_a}}^{{A_i}}(t)}, 0, {T_{{C_a}}^{{B_i} - }(t)c \varphi _{{C_a}}^{{A_i}}(t)}]$.
In the local coordinate system of the beam $i$, the force ${\bf{F}}_i^c(t)$ and moment $\textbf{M}_i^C(t)$ acting on the tip of the beam can be calculated using Eq.~\eqref{M8}.
\begin{equation}
\left\{ \begin{array}{l}
\textbf{M}_i^C(t) = \left( {{{\bf{P}}_c(t)} - {{\bf{P}}_i^b}(t)} \right) \times {{\bf{F}}^c}(t),\\
{\bf{F}}_i^c(t) = \prod\limits_{i}^{{N_c}} {{{\left( {{\bf{Ry}}({\theta _i}(t))} \right)}^{ - 1}}} {{\bf{F}}^c(t)},
\end{array} \right.
\label{M8}
\end{equation}
where $N_c$ is the joint number to which the force is applied. ${\bf{Ry}}$ is the rotation matrix around the y-axis. ${{\bf{P}}_c}(t)$ and ${{\bf{P}}_i^b}(t)$ is the location of the contact force and the beam $i$ tip in the global coordinate system. ${{\bf{P}}_i^b}(t)$ can be expressed by:
\begin{equation}
\begin{split}
{{\bf{T}}_i}(t) =\prod\limits_{i = 1}^{{i}} {\bf{Trz}}(L_c^i){{\bf{Tr}}({r_i},0,{z_i}){\bf{Ry}}({\theta _i})},\\
\end{split}
\label{M9}
\end{equation}
\begin{equation}
\left[ {\begin{array}{*{20}{c}}
{{\bf{R}}_i^b(t)}&{{\bf{P}}_i^b(t)}\\
0&1
\end{array}} \right] = {{\bf{T}}_i}(t),
\label{M9_1}
\end{equation}
where {${\bf{Tr}}$ is the translation matrix. {To further reduce the number of unknowns, assume that the contact point is at a distance $s_c$ from the proximal end along the robot backbone. ${{\bf{P}}_c}(t)$ can be expressed by Eq.~\eqref{M10},
{\small
\begin{equation}
\left\{ \begin{array}{l}
{{\bf{T}}_c}(t) ={{\bf{T}}_{N_c}}(t){\bf{Trz}}(s_{N_c}) {{\bf{Ry}}(\theta_c)}{{\bf{Ry}}(\frac{\pi}{2})}{\bf{Trz}}(\frac{d_c}{2}) ,\\
s_{N_c}=s_c-\sum_{i=1}^{Nc-1}(L_c^i+L_b^i)-L_b^{N_c},\\
\left[ {\begin{array}{*{20}{c}}
{{\bf{R}}_c(t)}&{{\bf{P}}_c(t)}\\
0&1
\end{array}} \right] = {{\bf{T}}_c}(t),
\end{array} \right. 
\label{M10}
\end{equation}
}
where $L_c^i$ is the length of each cable channel. $\theta_c$ is the deflection angle of the contact force on the plane perpendicular to the tangent of the backbone. $\theta_c=atan2(F^C_{yi}, F^C_{xi})$. {${\bf{Trz}}$ is the translation matrix along the z-axis.}}
\subsection{Beam Deformation in Local Coordinate System}
$r_i$, $z_i$, and $\theta_i$ can be calculated using the Bernoulli-Euler beam model.
{\small
\begin{equation}
E_i(t)I\frac{{{d^2}{\theta _i}(s,t)}}{{d{s^2}}} = {F_{zi}(t)}\sin {\theta _i}(s,t) - {F_{ri}(t)}\cos {\theta _i}(s,t),
\label{B1}
\end{equation}
}
{where $s$ represents the position along the beam measured from its base. The boundary can be expressed by the following:
\begin{equation}
\label{B2}
\left\{ \begin{array}{l}
\text{BC1}:{\theta _i}(0,t) = 0,\\
\text{BC2}:\frac{{d{\theta _i}(L,t)}}{{ds}} = \frac{{{M_{yi}}}(t)}{{{E_i(t)}{I_i}}},\\
\text{BC3}:{E_i}(t) = {E_A}\xi _i^A(t) + {E_M}(1 - \xi _i^A(t)),\\
\text{BC4}: {L_{C_a}^c}(t) = {L_{C_a}^{ct}}(t) + {L_{C_a}^e}(t),
\end{array} \right.
\end{equation} 
where BC1 and BC2 describe the mechanical nonlinearity of the beam, BC3 describe the material nonlinearity~\cite{zeng2021modeling}, and BC4 describe the cable length limitation. $\xi _i$ is the composition ratio of austenite inside the nickel-titanium skeleton, and the law of its composition change and bending angle is directly calibrated by reverse calibration. $L_{C_a}^{c}(t)$ is the input cable length in the robot. $L_{C_a}^e(t)$ is the length of elastic extension. $L_{C_a}^{ct}(t)$ is the total cable length of the cable in the continuum robot. }
$L_{C_a}^{ct}(t)$ and $L_{C_a}^e(t)$ can  be calculated using:
\begin{equation}
\label{B3}
\left\{ \begin{array}{l}
{L_{{C_a}}^{ct}(t)} = \sum\limits_{i=1}^{N_b} ({l_{C_a}^i(t) + {L_c^i}}), \\
{L_{C_a}^e(t)} = \frac{{{F_{{C_a}}^{in}(t)}{L_{os}}}}{{{E_{{C_a}}}{I_{{C_a}}}}},
\end{array} \right.
\end{equation} 
where $L_{os}$ is the unloaded length of the cables. 
Using the first two constraints, the Bernoulli-Euler beam can be simplified using the beam constraint model~\cite{zhang2024composite, wu2024body}:
{\small
\begin{equation}
\left\{ \begin{array}{l}
{{\bf{q}}_{i}} = \left[ {\begin{array}{*{20}{c}}
{\mathop {\hat{r_i}(t)}}\\
{\mathop {{\theta _i}(t)}}
\end{array}} \right] = {(A + \frac{{{L^2}{F_{zi}(t)}}}{{{E_i}(t)I_i}}B)^{ - 1}}\left[ {\begin{array}{*{20}{c}}
{\frac{{{L^2}{F_{ri}(t)}}}{{{E_i(t)}I_i}}}\\
{\frac{{{M_i(t)}L}}{{{E_i(t)}I_i}}}
\end{array}} \right],\\
\mathop {\hat{z_i}(t)} = \frac{{{a^2}{L^2}{F_{zi}(t)}}}{{12{E_i}(t)I_i}} + {\bf{q}}_{i}^{ - 1}\textbf{C}{{\bf{q}}_{i}} - \frac{{{L^2}{F_{zi}(t)}}}{{{E_i(t)}I_i}}{\bf{q}}_{i}^{ - 1}\textbf{D}{{\bf{q}}_{i}}+1,
\end{array} \right.
\label{B5}
\end{equation} }

where,
\begin{equation}
\begin{aligned}
& \mathbf{A}=\left[\begin{array}{cc}
12 & -6 \\
-6 & 4
\end{array}\right], \quad \mathbf{B}=\left[\begin{array}{cc}
1.2 & -0.1 \\
-0.1 & 0.13
\end{array}\right],  \\
&\mathbf{C}=\left[\begin{array}{cc}
-0.6 & 0.05 \\
0.05 & -0.067
\end{array}\right], \mathbf{D}=\left[\begin{array}{cc}
\frac{1}{700} & \frac{-1}{1400} \\
\frac{-1}{1400} & \frac{11}{6300}
\end{array}\right] ,
\end{aligned}
\label{B6}
\end{equation}
where $\hat{r_i}$, $\hat{z_i}$ are the dimensionless forms of the unknowns $r_i$, $z_i$ in the beam deformation equation, 
\begin{equation}
\left\{ \begin{array}{l}
{z_i}(t) =  \hat{z}_i(t)L_{b}^i,\\
{r_i}(t) = \hat{r}_i(t)L_{b}^i.\\
\end{array} \right.
\label{B7}
\end{equation}

\subsection{Geometric Constraints and Optimization} 
\renewcommand{\algorithmicrequire}{ \textbf{Input:}} 
\renewcommand{\algorithmicensure}{ \textbf{Output:}} 
\begin{algorithm}[t]

\caption{Shape, contact force, and contact location sensing for cable-driven continuum robots}
\label{ALG1}
\begin{algorithmic}[1]

\REQUIRE{Set cable length $L_{C_a}^s(t)$, input cable forces ${\bf{F}}_{C_a}^{in}(t)$, forces information ${{\bf{F}}^M} (t)$ from F/T sensor.}
\ENSURE{Estimated contact force ${{\bf{F}}^C} (t)$, estimated contact location $s_c$, shape information $\textbf{P}^b_i(t)$}
\STATE {Initialize:} $k=1,z_i(0,t)=l^i_b,x_i(0,t)=0,\theta_i(0,t)=0$.
\STATE Calculate the contact force in the world coordinate system according to Eq. \eqref{F1}.
\FOR {For each $s_c$.}
\WHILE{$\theta_i(k+1,t)-\theta_i(k,t)<\epsilon$}
\STATE Calculate the contact force in the local coordinate system according to Eq. \eqref{M8}.
\STATE Solve for the forces and moments acting on the beam elements inside the robot according to Eq. \eqref{M7}.
\STATE Solve BC1 and BC2 according to Eq. \eqref{B5} and Update beam deformation parameters $z_i(k+1,t),x_i(k+1,t),\theta_i(k+1,t)$ in the local coordinate system.
\STATE Solve for the coordinates of the robot backbone in the world coordinate system according to Eq. \eqref{M9_1}.
\STATE Update the $E_i$ according to BC3, $k=k+1$.
\ENDWHILE
\ENDFOR
\STATE Calculate the cable length $L_{{C_a}}^c(t)$ including the variable $s_c$ according to Eq. \eqref{B3}.
\STATE Calculate cost function $\Delta L_{{C_a}}^c({\bf{x}})$ according to Eq. \eqref{B10}.
\STATE Solve BC4 according to Eq. \eqref{B9}.
\STATE Extract shape information from the optimization process.
\end{algorithmic}
\end{algorithm}
\vspace{-1mm}

Using the constraints, the above is transformed into the form of an optimization function.
\begin{equation}
\begin{array}{c}
{\bf{x}} = s_c,\\
\mathop {{\rm{minimize}}}\limits_{\bf{x}}( \sum\limits_{a = 1}^{{N_b}} {{{\left\| {\Delta L_{{C_a}}^c({\bf{x}})} \right\|}^2}  } ),\\
{\rm{s.t.}}F_{{C_a}}^{in} \ge 0{\text{ ,when, }}F_{{C_a}}^{in} = 0,{\rm{ }}\Delta L_{{C_a}}^c({\bf{x}}) = 0.
\end{array}
\label{B9}
\end{equation}
where
\begin{equation}
\Delta L_{{C_a}}^c({\bf{x}}) = L_{{C_a}}^s({t}) - L_{{C_a}}^c(t).\\
\label{B10}
\end{equation}
The objective function minimizes the discrepancy between the set cable length $L_{C_a}^s(t)$ and the cable length $L_{C_a}^c(\mathbf{x})$ calculated from the contact position. The inequality constraint ensures that the cable force $F_{C_a}^{in}$ remains non-negative, as cables can only sustain tension. }
Solving the above equations can realize the proprioception (shape, contact force, and contact location) of the robot, as shown in Algorithm \ref{ALG1}.

\section{Results}
\subsection{Device}

 \begin{figure*}[t]
\centering  
\includegraphics[width=180mm]{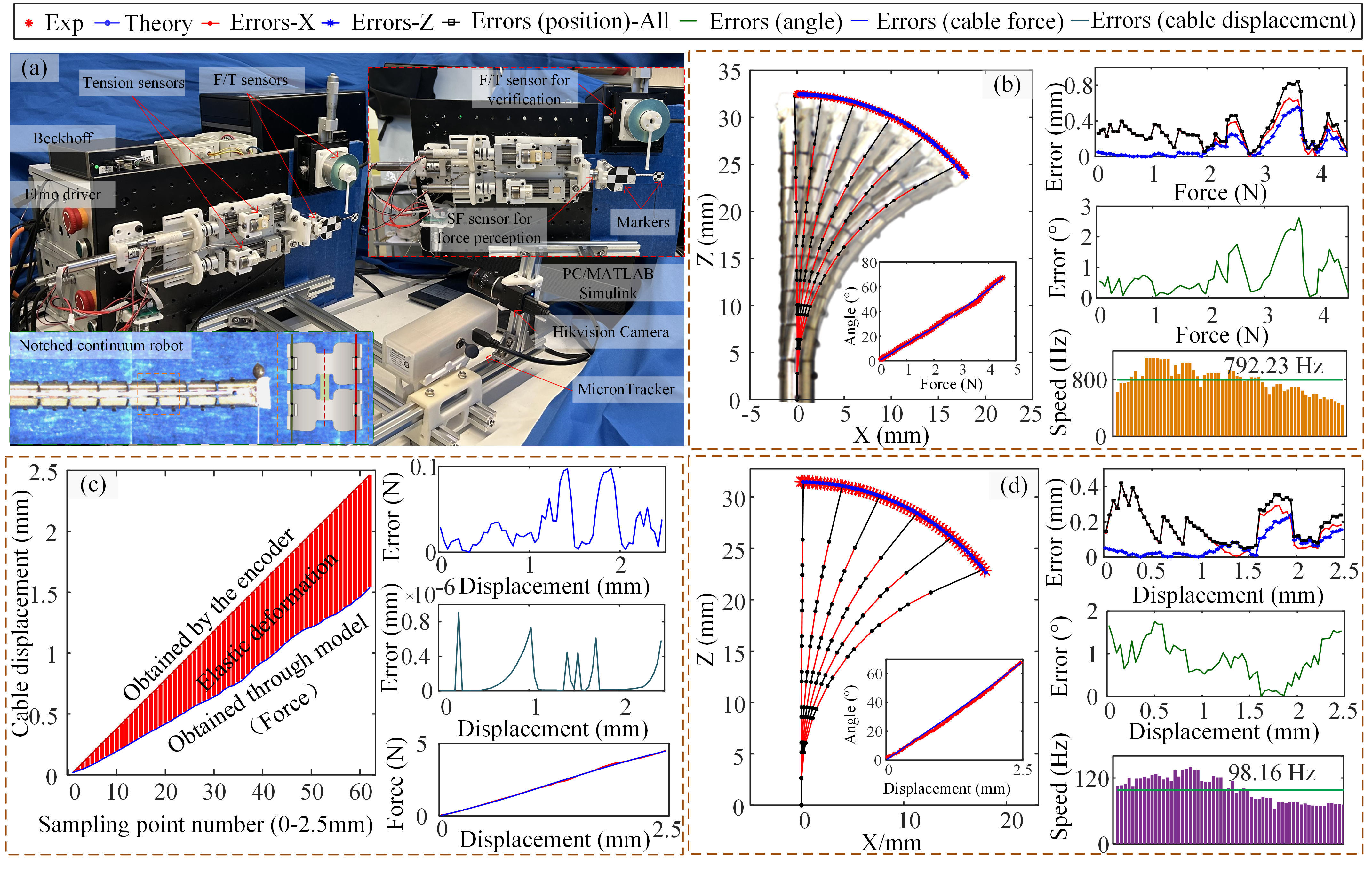}
\caption{Test benches. (a) Experimental platform. (b) Shape prediction of the robot under force control. (c) Elastic elongation of the driving cable and elastic deformation of the component. (d) Shape prediction of the robot under displacement control.}
\label{Ex1}   
\vspace{-3mm}
\end{figure*}

{The device shown in Fig.~\ref{Ex1}(a) is used to verify the effectiveness of the proposed method. Two motors (Re16, Maxon, Switzerland) drive two cables (Diameter: 0.27 mm, TESAC, Japan) loaded with tension sensors (MLP, Transducer Techniques, USA, range: 10 pounds, resolution: 1/100 pounds, sampling rate: 1600 Hz) to control the bending degree of freedom on a plane. One motor controls the robot's forward movement.
An F/T sensor (Nano 17, ATI, USA, range: Fx and Fy:12 N, Fz:17 N, Tx, Ty, Tz: 120 N·mm, resolution: 1/320 N for force and 1/64 N·mm for torque, sampling rate: 7000 Hz) is fixed on the proximal end of the robot to collect the six-dimensional force information, and another F/T sensor (TFS-HKM60-N10, BinLi, China, range: Fx, Fy and Fz:10 N, Tx, Ty, Tz: 100 N·mm, resolution: 1/100 N for force and 1/10 N·mm for torque, sampling rate: 2000 Hz) is fixed on a linear moving platform to apply contact force. 
Two markers are fixed on the base and tip of the robot, respectively. The robot's motion trajectory information is collected using a binocular camera (MicronTracker, Claron Technology, Canada). Another camera (CS060, Hikrobot, China) collects the shape information.
A notched continuum robot (NCR) is used for verification. The robot's structure is shown in Fig.~\ref{Ex1}(a), and its parameters are shown in Table~\ref{NCR}. $D_o$, $D_i$, and $w$ represent the robot's outer and inner diameters, and the beam width, respectively.} 

\begin{table}[t]
	\centering
	\caption{Structural parameters of four NCR (mm)}
    \resizebox{0.48\textwidth}{!}{	
	\begin{tabular}{cccccccccccc}
		\hline
		 $D_o$ & $D_i$ & $d_c$ & $w$ & $h$ & $L^i_{C_a}$& $L^i_{c}$ & $N_c$ & $E$ & $E_m$ & $\mu$ \\ 
         \hline
        3.5 & 3.2 & 2.6 & 0.4 & 2.5 &2.1 & 0.9 & 7&47.05~Gpa&0.08 Gpa&0.33\\ 
        \hline      
	\end{tabular}
    }
	\label{NCR}
\end{table}


\subsection{Shape Estimation}
\subsubsection{Shape Estimation Under Force Constraints}

A continuous displacement of 0–2.5 mm was slowly applied to the robot’s drive cable in the absence of external contact force, and the corresponding dynamic cable tension and tip position were recorded to validate the accuracy of the model. Then, the displacement of the driving cable was changed by 0.5 mm intervals, and the shapes of the theoretical and actual robots were recorded, as shown in Fig.~\ref{Ex1}(b). From the figure, we can see that the theoretical shape and actual shape, the theoretical trajectory and actual trajectory of the robot are almost completely overlapped. {The average tip position error of the robot is 0.41 mm, the maximum tip position error is 0.83 mm, the average tip deflection angle error is 1.17°, and the maximum tip deflection angle error is 2.77°. These errors and transient error peaks mainly stem from slight delays in the external visual measurement system and slight stutters caused by uneven friction during movement.} The average computational efficiency under force constraints in shape evaluation is 792.23 Hz in MATLAB. 

\subsubsection{Shape Estimation with Displacement Constraints}
{Before shape estimation under displacement control, a calibration was performed to account for the elastic elongation of the drive cables and the compliance of the transmission components. The difference between the measured and the kinematically predicted cable displacement was modeled using a piecewise spline function (Fig.~\ref{Ex1}(c)). With this compensation active, the model’s ability to predict the internal cable tension from a prescribed displacement was evaluated. The average error between the model-predicted and the sensor‑measured cable tension was 0.046 N, while the residual displacement error remained below $10^{-3}$ mm. Subsequently, the overall shape‑estimation performance was tested by applying cable displacements of 0–2.5 mm without external contact. The reconstructed shape and tip trajectory closely match the experimentally observed ones (Fig.~\ref{Ex1}(d)). Quantitatively, the average tip‑position error was 0.29 mm (max 0.45 mm) and the average tip‑orientation error was 0.84° (max 1.81°). The average computational rate for the complete shape‑estimation routine reached 98.16 Hz in MATLAB.}

\begin{figure*}
\centering  
\includegraphics[width=180mm]{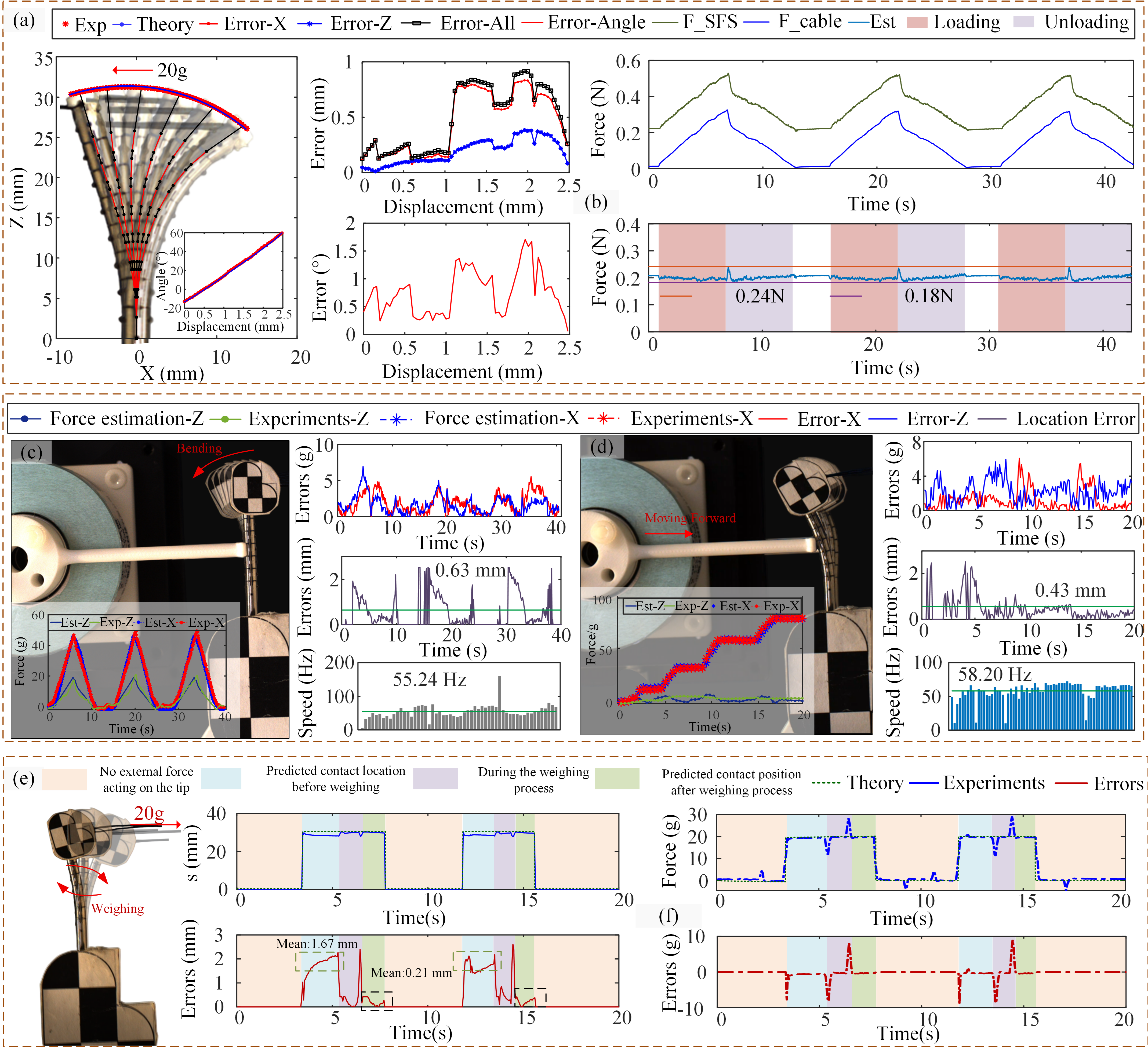}
\caption{Force prediction test results. (a) Robot shape under tip load. (b) Estimation of tip load. (c) Contact force and contact point prediction for active contact. (d) Contact force and contact point prediction for passive contact. (e) Reducing perception errors through repeated weighing movements. (f) Improvement of contact location estimation through friction redistribution.
}
\label{Ex2}  
\vspace{-2mm}
\end{figure*}

\subsection{Tip Contact Force Perception}

A known contact force was applied to the robot tip to test its shape estimation capability. The results are shown in Fig.~\ref{Ex2}(a). Under the cable displacement constraint, the theoretical shape of the robot and the image of the actual robot basically coincide. A continuous displacement of 0 to 2.5 mm was slowly applied to the drive cable, and the theoretical and actual tip displacement and deflection angle were recorded.
The model's prediction error for the robot's average tip position is 0.56 mm, with a maximum displacement error of 0.97 mm. In addition, the average error in the deflection angle of the robot tip is 0.94°, while the maximum error is 1.56°.
We can easily calculate the dynamic contact force of the robot tip based on the information of the F/T sensor and the tension sensor using the method in the previous section. The calculation results are shown in Fig.~\ref{Ex2}(b). The average error of the dynamic tip contact force perception during the loading process is 0.47 g, and the max estimated error is 1.48 g. The average error of the tip contact force perception during the entire movement process is 0.93 g, and the max error is 4.02 g.

\subsection{Body Contact Force and Location Perception}

{Here, we perceive the magnitude of the contact force and the position of the point of action for the two types of contact mentioned above. To achieve stable and repeatable force loading on the curved surface of a continuum robot, a ring contactor was used in the experiment. When the contactor contacts the inclined robot surface and undergoes local deformation, the two adhere along a stable contact line, effectively preventing relative slippage. Within the robot's bending plane (i.e., the two-dimensional observation plane), this contact line projects as an equivalent contact point, thus maintaining consistency with the point contact assumption of the model in the plane in terms of mechanical effects.}
\subsubsection{Contact Force perception under active body contact}
First, we used the continuum robot to actively collide with obstacles and used an additional F/T sensor to perceive the force of active contact. The experimental results are shown in Fig.~\ref{Ex2}(c). The average force prediction errors of the robot in the X and Z directions are 1.82 g and 1.64 g, respectively. The average contact point position prediction error along the robot backbone direction is 0.63 mm. {The average computational efficiency for estimating body force is 55.24 Hz in MATLAB.}

\subsubsection{Contact Force perception under passive body contact}
Similarly, the body of the continuum robot was touched by an obstacle to enable the robot to perceive passive contact. The experimental results are shown in Fig.~\ref{Ex2}(d). The average force prediction errors of the robot in the X and Z directions are 1.23 g and 2.20 g, respectively. The average contact point position prediction error along the robot backbone direction is 0.43 mm. {The average computational efficiency for estimating body force under passive contact is 58.20 Hz in MATLAB. Contact position errors mainly arise from two sources: force estimation inaccuracies propagating through the optimization framework, and minimal contact interface slippage that may occur despite the ring contactor design. }
 
\subsection{Force Perception Strategy Inspired by Human Behavior}

When a contact force acts on the continuum robot, the robot undergoes passive deformation, which disrupts the initial friction distribution and leads to an inaccurate estimation of the force application position. To address this issue, we introduce a recalibration procedure analogous to how a person repeatedly hefts a heavy object to adjust to its weight. Specifically, once the robot is subjected to an external load, it is driven to perform a short reciprocating motion. This process redistributes the cable friction into a stable state, thereby restoring the accuracy of contact force perception.

The experimental results are shown in Fig.~\ref{Ex2}(e). After the reciprocating motion, the average error of the estimated force application position is reduced from 1.67 mm to 0.21 mm.

\begin{figure*}
\centering  
\includegraphics[width=180mm]{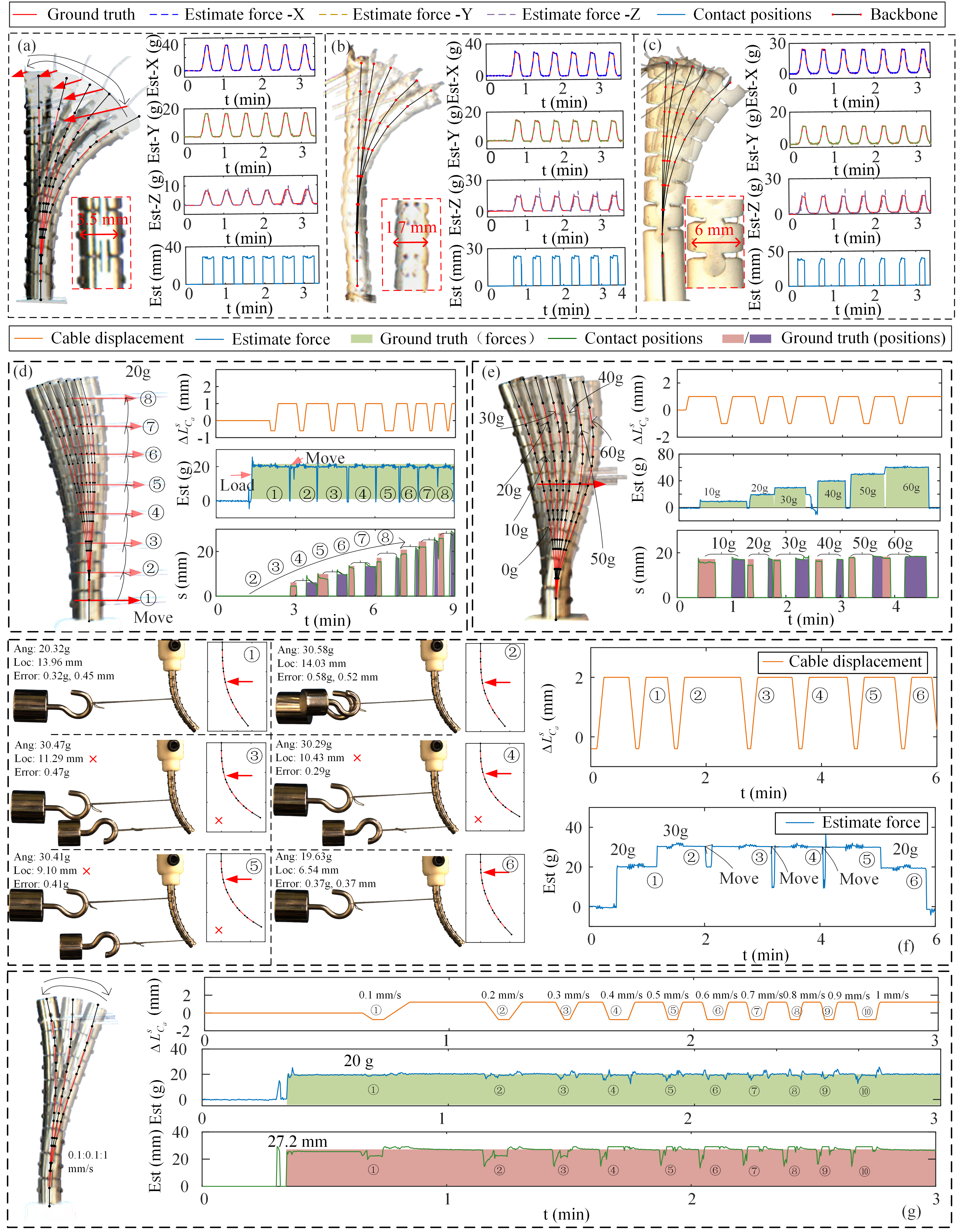}
\caption{Three-dimensional force estimation and stability verification. (a)–(c) Three-dimensional force estimation of continuum robots with different diameters. (d)-(e) Perception framework validation under different contact conditions. (f) The performance of the perception framework in multi-point contact. (g) Performance of the perception framework at different speeds.
}
\label{Ex3}  
\vspace{-2mm}
\end{figure*}

\subsection{Three-dimensional force perception test and universality test}
\subsubsection{Three-dimensional force perception test} {To validate the proposed framework under multi-axis loading conditions, a 3D force perception test was conducted using the setup illustrated in Fig. 2(a). The reference F/T sensor was positioned 20 mm lower than the robot base, establishing a deliberate 3D angular offset with the robot. A rubber band was attached between the robot’s tip and the F/T sensor.} {The robot was then actuated through prescribed cable displacements to generate a range of three-dimensional contact forces. Force estimation was performed on a notched-tube continuum robot with a diameter of 3.5 mm. As shown in Fig.~\ref{Ex3}(a), the average force estimation errors along the X-, Y-, and Z-axes are 0.96 g, 0.60 g, and 0.88 g, respectively. The average contact position estimation error is 0.73 mm.}

\subsubsection{Universality test}
{To evaluate the general applicability of the framework, experiments were extended to two additional continuum robot designs with distinct mechanical structures and scales.
First, the framework was applied to a miniature endoscopic snake-bone continuum robot with a diameter of 1.7 mm. The results, presented in Fig.~\ref{Ex3}(b), show average force estimation errors of 0.49 g, 0.31 g, and 0.80 g for the X-, Y-, and Z-axes, respectively, with an average contact position error of 0.61 mm.
Subsequently, the framework was tested on a larger multi-vertebra continuum robot with a diameter of 6 mm. As depicted in Fig.~\ref{Ex3}(c), the average force estimation errors are 0.69 g, 0.47 g, and 0.91 g for the three axes, while the average contact position error is 1.35 mm.
}

\subsection{Reliability and stability testing}

\subsubsection{Validation of the perception framework at different contact positions.}
To evaluate the robustness of the proposed sensing framework to different contact positions, a 20 g weight was suspended via a thin wire at several locations along the robot body. The robot was then moved repeatedly while the contact force and location were estimated using the proposed proprioception method. Results are summarized in Fig.~\ref{Ex3}(d).
The accuracy of force magnitude estimation remained consistently high across all tested positions, with average errors below 1.5 g, indicating that the force‑decoupling strategy is largely independent of contact location.
In contrast, contact‑location estimation showed a dependence on the position. The average error decreased from 2.11 mm for a proximal contact to 0.64 mm for a distal (near‑tip) contact. This trend can be attributed to the larger structural deformation induced by an external load when it is applied closer to the tip, which improves the sensitivity of the shape‑based localization in the optimization framework.

\subsubsection{Validation of the perception framework at different contact force magnitudes.}
To assess the performance of the proposed framework under varying contact forces, weights ranging from 0 g to 60 g were suspended at the same mid‑body position in 10 g increments. The results are presented in Fig.~\ref{Ex3}(e). The force magnitude estimation remained highly accurate across all loading levels, with average errors consistently below 1.5 g. This demonstrates the robustness of the force‑decoupling process over a clinically relevant force range. In parallel, the contact‑location estimation showed a clear trend: as the applied force increased, the average localization error decreased from 1.67 mm at 10 g to 0.49 mm at 60 g. This improvement can be attributed to the larger structural deformation induced by higher forces, which provides a more pronounced mechanical signature for the optimization‑based localization algorithm to resolve.

\subsubsection{The performance of the perception framework in multi-point contact.}
To investigate the behavior of the proposed framework when multiple contact points are present, we designed six loading scenarios: (1) a single 20 g weight suspended at the mid‑body; (2) an additional 10 g weight added to the same location (total 30 g); (3) 10 g and 20 g weights placed at two distinct locations (total 30 g distributed); (4) the single 20 g weight moved to a different position; (5) the single 20 g weight moved to a different position; and (6) only the 20 g weight reapplied at its original location.
Results are summarized in Fig.~\ref{Ex3}(f). The framework consistently estimates the resultant force of the multiple contacts with high accuracy (error $<$ 1.5 g), demonstrating that the force‑decoupling process remains effective even under distributed loading. However, as the mechanical model assumes a single point contact, the shape reconstruction and contact‑location estimation converge to an equivalent single point that minimizes the cable‑length residual in the optimization. Consequently, the reconstructed shape and the estimated contact position deviate from the true multi‑point configuration.

\subsubsection{Performance of the perception framework at different speeds.}

To investigate the influence of motion dynamics on perception accuracy, a 20 g weight was fixed to the robot tip and the cable was displaced at speeds ranging from 0.1 mm/s to 1.0 mm/s, as shown in Fig.~\ref{Ex3}(g). During the motion phase, the estimated contact force exhibited a systematic transient variation: it increased during acceleration and decreased during deceleration, with the amplitude of variation slightly increasing at higher speeds. This transient behavior originates from the inertial effect of the reference weight.
After the robot came to rest, both the magnitude and location of the contact force were estimated with high accuracy: the average force error remained below 1 g, and the average contact‑position error was under 1 mm. These results confirm that the proposed sensing framework effectively captures dynamic force and can accurately estimate static position once motion stabilizes.

\section{Discussion}
In this paper, the perception of contact forces is implemented in two sequential steps. In the first step, cable tension measurements and data from an F/T sensor are combined to estimate the dynamic contact force of the cable-driven continuum robot. Under the constraint of constant cable length, this strategy effectively decouples the complex interactions among contact force, proximal end forces, and cable tensions, enabling accurate estimation of contact force. In the second step, the contact location is inferred using a high-fidelity static model of the continuum robot based on the contact force sensed in the first step. This method avoids solving a multivariate nonlinear equation system involving variables such as contact force and contact location, which can reduce computational complexity and allow real-time perception (Over 50 Hz in MATLAB).

{The proposed method demonstrates inherent robustness through its design architecture. The dual sensing channels—cable tension and base force measurements—provide mutually constrained information through the force-balance relation (Eq. 1), which helps to identify and reject common-mode sensor errors arising from factors like thermal drift or mechanical vibration. Furthermore, formulating contact localization as a single-variable optimization problem (Eq. 21) avoids the ill-conditioning often encountered in multi-dimensional estimation approaches. Together, this complementary sensing strategy and regularized optimization framework collectively suppress the influence of sensor noise and model uncertainties. Experimental results confirm stable performance under various dynamic conditions, including friction changes and reciprocating motion (Section III.E).}

{
A unique advantage of this method lies in its ability to reveal physical phenomena that are not explicitly captured by the static model. As shown in Fig.~\ref{Ex2}(e), the combined perception strategy effectively captures sudden transitions in dynamic and static friction, as well as the effects of inertial forces due to object acceleration—even though these factors are not explicitly modeled. The robustness of the proposed strategy primarily stems from its reliance on dynamic measurements from the F/T sensor and cable tension sensors, rather than on static or quasi-static assumptions, thereby enabling accurate estimation of contact force magnitude under varying motion states. This robustness is particularly promising, as it demonstrates that the proposed strategy possesses a degree of universality and can adapt to a wide range of practical conditions.}

In comparison to existing strategies, the proposed method offers several distinct advantages. First, relative to conventional model-based approaches—which typically require iterative solutions of complex nonlinear equations with considerable computational overhead—our method solves the boundary-value perception problem through an efficient single-variable optimization. This not only enhances computational efficiency, but also extends the framework’s capability to multi-dimensional force sensing and contact localization, a task difficult to achieve with purely analytical models under unknown contact conditions.
Second, compared to sensor-based methods that rely on distributed sensing elements along the robot body (e.g., FBG arrays or soft strain sensors), our approach requires only base-mounted sensing—a six-axis F/T sensor and cable tension feedback. This configuration greatly simplifies system integration, preserves the structural and functional integrity of the slender robotic shaft, and avoids occupying the limited cross-sectional space in submillimeter surgical instruments, while still enabling simultaneous estimation of contact forces, localization of interaction points, and reconstruction of the robot’s shape.
Third, unlike previously reported hybrid approaches that combine localized embedded sensing with mechanical models, our framework employs a proximal-sensing-only paradigm integrated with model-based optimization. This design avoids physical interference with the robot’s constrained workspace, eliminates dependencies on external tracking or vision systems, and reduces the computational complexity typically associated with distributed sensor fusion in hybrid frameworks.
Moreover, compared to observer-based, learning-enhanced hybrid methods, and mode-based methods that often presume known contact locations, our method actively resolves the underdetermined nature of force and shape estimation in unknown, contact scenarios—enabling reliable proprioception under clinical uncertainties without sacrificing real-time performance.

{
It is also worth noting that the proposed sensing framework is applicable to continuum robots of varying diameters, as the overall operation of the framework does not depend on structural diameters. For multi-segment continuum robots, the computational efficiency remains largely unaffected because the optimization equations consistently involve only one unknown $s_c$; increasing the number of segments does not increase the dimensionality of the optimization problem.}

{

A noteworthy observation from Experiment C concerns the transient accuracy during directional reversal. The force estimation error averaged 0.47 g during steady loading phases, compared to 0.93 g over the complete motion cycle. This difference is attributable to inertial transients during acceleration/deceleration at direction reversals.}

{The proposed framework is designed and validated for single-point contact scenarios, estimating only a single resultant force vector and an equivalent contact location. Under simultaneous multi-point contact, the optimization converges to an equivalent single contact location and resultant force, which does not represent the true distributed interaction and introduces errors in shape reconstruction and force attribution. Extending the framework to accommodate multi-point or distributed loading—through reformulated optimization or integrated distributed sensing—remains a crucial direction for future work.}

\section{Conclusion}
In this paper, a novel proprioception method for micro-scale cable-driven continuum robots is presented, addressing the challenges of multi-dimensional contact force and contact location perception in minimally invasive surgical environments. By combining proximal end measurements from an F/T sensor with internal cable tension data and integrating these with a nonlinear physical model, the proposed approach achieves accurate three-dimensional force estimation and shape reconstruction along the robot body. This framework is inspired by the tendon-joint collaborative sensing mechanism of the human finger, and a quasi-bionic mapping was established to transfer integrated sensing strategies from human tissues and joints to the robotic system.

Several key advantages of the proposed method can be highlighted. First, the approach does not require sensors along the robot body, eliminating constraints on robot size and enabling application to continuum robots of varying diameters. Second, by leveraging proximal measurements as conditions, the method resolves the underdetermined nature of force and shape estimation, allowing stable perception without relying on quasi-static assumptions. Third, the optimization framework effectively decouples the complex interactions among mechanical and material nonlinearities, robot motion states, and contact forces, enabling real-time estimation and comprehensive intraoperative awareness. Compared with purely model-based, sensor-based, or hybrid approaches, this method reduces computational complexity, simplifies sensor integration, and avoids occupying the robot’s limited workspace, while still achieving force and shape perception.

Future work will focus on applying this method to a surgical robot for suturing, aiming to provide surgeons with contact force feedback through haptic teleoperation devices.

\bibliographystyle{ieeetr}
\bibliography{Reference1}

\end{document}